\pdfoutput=1

\documentclass[11pt]{article}

\usepackage[preprint]{acl}

\usepackage{times}
\usepackage{latexsym}

\usepackage[T1]{fontenc}

\usepackage[utf8]{inputenc}

\usepackage{microtype}

\usepackage{inconsolata}

\usepackage{graphicx}

\usepackage{url}
\usepackage{booktabs}
\usepackage{multirow}
\usepackage{subcaption}
\usepackage{bbding}
\usepackage[most]{tcolorbox}

\newcommand{\MNB}{\textsc{MNB}}
\newcommand{\MVB}{\textsc{MVB}}
\newcommand{\SVC}{\textsc{SVM}}
\newcommand{\BoW}{\textsc{BoW}}
\newcommand{\LSA}{\textsc{LSA}}
\newcommand{\GloVe}{\textsc{GloVe}}
\newcommand{\Human}{\textsc{Human}}
\newcommand{\Majority}{\textsc{MajorityClass}}

\newcommand{\BERT}{\textsc{BERT}}
\newcommand{\DistilBERT}{\textsc{DistilBERT}}
\newcommand{\GPT}{\textsc{GPT-2}}
\newcommand{\Llama}{\textsc{Llama3.1-Instruct}}
\newcommand{\BiomedBERT}{\textsc{BiomedBERT}}
\newcommand{\ChatGPT}{\textsc{GPT-4o-mini}}
\newcommand{\HYPE}{\texttt{Hype}}
\newcommand{\NOTHYPE}{\texttt{Not Hype}}

\title{Hype or not? Formalizing Automatic Promotional Language Detection in Biomedical Research}

\author{
 \textbf{Bojan Batalo\textsuperscript{1}},
 \textbf{Erica K. Shimomoto\textsuperscript{1}},
 \textbf{Neil Millar\textsuperscript{2}},
\\
 \textsuperscript{1}National Institute of Advanced Industrial Science and Technology (AIST),
 \textsuperscript{2}University of Tsukuba,
\\
 \texttt{\{bojan.batalo,kidoshimomoto.e\}@aist.go.jp, millar.neil.gm@u.tsukuba.ac.jp}
}

\begin{document}
\maketitle
\begin{abstract}

In science, promotional language ('hype') is increasing and can undermine objective evaluation of evidence, impede research development, and erode trust in science. In this paper, we introduce the task of automatic detection of hype, which we define as hyperbolic or subjective language that authors use to glamorize, promote, embellish, or exaggerate aspects of their research. We propose formalized guidelines for identifying hype language and apply them to annotate a portion of the National Institutes of Health (NIH) grant application corpus. We then evaluate traditional text classifiers and language models on this task, comparing their performance with a human baseline. Our experiments show that formalizing annotation guidelines can help humans reliably annotate candidate hype adjectives and that using our annotated dataset to train machine learning models yields promising results. Our findings highlight the linguistic complexity of the task, and the potential need for domain knowledge and temporal awareness of the facts. While some linguistic works address hype detection, to the best of our knowledge, we are the first to approach it as a natural language processing task.

\end{abstract}

\section{Introduction}

Detecting AI-generated content, plagiarism, fake news, and bias related to health topics has become the target of many machine learning systems to ensure evidence-grounded information is correctly communicated to the general audience. However, one issue yet to be addressed is the increasing use of promotional language – a phenomenon referred to as 'hype' \cite{millar2019important}.

For example, investigators promote the significance and novelty of their research using exaggerated terms (\textit{revolutionary}), dramatically describe research problems (\textit{daunting}), amplify the scale and rigor of their methods (\textit{extensive}, \textit{robust}) and the utility of the results (\textit{actionable}, \textit{impactful}). Increasing use of hype has been demonstrated in biomedical funding applications \cite{millar2022trends} and journal publications \cite{vinkers2015use}, while comparable trends are evident in other research fields \cite{weidmann2018use}.

\begin{figure}[t]
\centering
\includegraphics[width=\linewidth]{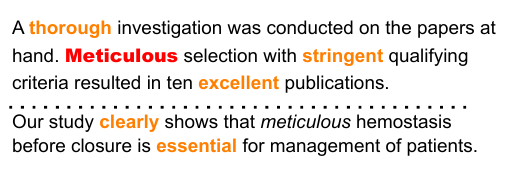}
\caption{Examples of 'hype' such as gratuitous amplification of research rigor; determining hype depends on the context, as exemplified by adjective 'meticulous'.}
\label{fig:sample}
\vspace{-6pt}
\end{figure}

Hype in science is a cause for concern. As the former editor-in-chief of JAMA Network journals points out, words such as \textit{groundbreaking}, \textit{transformative}, or \textit{unprecedented} are rarely justified and may undermine objective assessment, impeding the development of further studies, policies, clinical practice, and knowledge translation \cite{bauchner2023hype}. Moreover, promotional and confident language can bias readers' evaluation of research \cite{van2022effect, peng2024promotional}, and public trust in science is eroded when promotional language creates unrealistic expectations or misrepresents findings~\cite{intemann2022understanding}.

Previous linguistic studies have sought to detect hype language through collecting and analyzing raw corpora of biomedical research texts~\cite{millar2019important, millar2022trends, millar2023promotional}, and have identified a lexicon of 140 adjectives that can carry promotional meaning in biomedical texts ~\cite{millar2022trends,millar2022trendsfoa}. Although this lexicon can help identify candidates, discussions about whether a specific term constitutes hype remain problematic, as a given word or phrase may be promotional depending on the context. For instance, adjectives like \textit{essential} and \textit{meticulous} can promote significance or rigor, but they may also occur in a neutral context or technical phrase (e.g., essential fatty acid, meticulous hemostasis), as shown in Figure~\ref{fig:sample}. 

Therefore, in this work, we introduce the task of automatic detection of hype language, focusing on biomedical research texts. To the best of our knowledge, we are the first to address it as a natural language processing (NLP) task. As a starting point, and highlighting that this is not an exhaustive definition, we follow~\citet{millar2019important} and define the concept of hype as `hyperbolic and/or subjective language that authors use to glamorize, promote, embellish and/or exaggerate aspects of their research.' Likewise, we consider their lexicon of adjectives as a starting point, as adjectives are prototypical and most common means of expressing evaluation \cite{martin2003language}.

To solve this task, we first address data annotation. Determining whether a word is used with promotional intent involves subjective judgment based on context and interpretation. Moreover, the general definition of hype may be inadequate for distinguishing ambiguous cases, forcing annotators to rely on intuition and group discussion. Therefore, we propose formal annotation guidelines to determine whether an adjective is used in a promotional manner based on its semantics, function, and context. Using these guidelines, we then manually annotate a dataset of 550 sentences from the NIH grant application abstract corpus containing potentially promotional adjectives.

Finally, we formulate automatic hype detection as a text classification task, and test some traditional NLP classifiers using bag-of-words and word embeddings as features, as well as finetuning small scale language models, such as \BERT{} and \GPT{} and using off-the-shelf large language models such as \Llama{} and \ChatGPT. We compare their performance to a human baseline that had no knowledge of our guidelines. Our results indicate that formalizing annotation guidelines helps humans reliably annotate adjectives as \HYPE{} or \NOTHYPE, and training NLP models on the annotated dataset yields promising results for automatic detection of hype.

\section{Related Works}
\label{sec:related_works}

To the best of our knowledge, no prior work withing NLP has been done on promotional language use in scientific publishing. However, here we highlight some work relevant to the analysis of such language in various publication outlets.

\paragraph{Tackling promotional language} \citet{bhosale-etal-2013-detecting} propose a model for detecting promotional content in Wikipedia articles. They use both n-gram language models and probabilistic context-free grammars and show that incorporating stylistic and syntactic features outperforms models trained only on lexical and meta-features. 

In addition, several works examine exaggerated claims and statements related to health news. \citet{li-etal-2017-nlp} analyze the discrepancies in the strength of claims made in scientific journal articles, and news articles in which they are reported to the public, with the aim of detecting media manipulation; the authors use a bag-of-words approach with lexical features to train simple classifiers. \citet{patro-baruah-2021-simple} expand on the same task and dataset, and utilize BERT~\cite{devlinBERTPretrainingDeep2019} to extract relation phrases, used to train classifiers to recognize the strength of a claim made in the journal article and its corresponding news reporting, which are compared. A similar task is undertaken by \citet{wright-augenstein-2021-semi}, who analyze pairs of journal articles and their press releases, as well as by \citet{kamali-etal-2024-using}, who focus on persuasive writing for health disinformation. 

Works on exaggeration focus on the amplification of relational claims between an original source and the reporting text (X \textit{might cause} cancer vs. X \textit{causes} cancer). These studies primarily examine modifications of certainty and confidence (commonly referred to as epistemic stance) typically expressed through modal verbs (\textit{might}, \textit{will}, \textit{can}), hedges (\textit{likely}, \textit{clearly}), and causal triggers (\textit{causes}, \textit{leads to}). Our work on hype is primarily concerned with value judgments (\textit{innovative}, \textit{promising}, \textit{impactful}). While the intensity of hype can be modified by epistemic stance features, these are not central to its identification.

\paragraph{Understanding underlying intentions}
Simultaneously, there is a growing interest in understanding the underlying intentions behind utterances, exemplified by tasks such as sentiment analysis~\cite{zhang2023sentiment}, emotion analysis~\cite{del2024emotion}, and stance detection~\cite{mu2024examining}. Sentiment analysis, one of the most prominent NLP tasks, identifies whether a writer’s orientation towards an object, person, or idea is positive, neutral, or negative. Closely related to this is emotion analysis, which focuses on specific emotions, whereas the study of stance examines attitudes, judgments, and commitment. 

We find similarities between hype detection and sentiment analysis, especially in its fine-grained version, as words with similar meanings can vary in promotional intensity (e.g., \textit{new} vs. \textit{novel} vs. \textit{innovative} vs. \textit{groundbreaking}). However, while sentiment analysis deals with only two polarities of sentiment, i.e., positive and negative, hype language can be used to promote different aspects of research~\cite{millar2022trends}, and, therefore, it relates more closely to emotion detection. Also, just like emotion analysis and stance detection, hype detection is highly context-dependent; a given word or phrase may be promotional in one setting while not in others. 

Furthermore, the detection of hype language is tied to identifying authors' intention to promote aspects of their research. Such intention can be expressed through different choices of words, which connects our task to lexical choice~\cite{ding2021understanding}. This task aims at choosing words to communicate the intended information to the reader in a generated text.

Nevertheless, a key distinction of our task compared to the aforementioned approaches is that it requires domain knowledge of the specific research field — in this case, biomedical research. In most cases, the use of words such as \textit{groundbreaking} or \textit{revolutionary} are rarely justified~\cite{bauchner2023hype}. However, if we look at the history of medicines, for example, the development of penicillin was revolutionary since, before its introduction, there was no effective treatment for infections such as pneumonia, gonorrhea, or rheumatic fever. 

\section{NIH grant application abstract corpus}
\label{sec:data}

The starting point of our work is the corpus of raw texts compiled by~\citet{millar2022trends}, which comprises 901,717 abstracts of successfully funded grant applications submitted to and approved by the National Institutes of Health (NIH) in the United States of America. The NIH corpus has 335 million words, out of which 36.4 million are adjectives. \citet{millar2022trends} give the following broad definition of hype: \textit{``If the adjective has \textbf{positive value judgment}, and can be \textbf{removed or replaced without loss in meaning}, it is potentially hype''}. Following this definition, the authors have identified 140 adjectives as 'potentially hype' in the NIH corpus. 

These adjectives were categorized in eight groups, based on the aspects of research  they are promoting: 
\textbf{importance} (\textit{critical}, \textit{ fundamental}), \textbf{novelty} (\textit{innovative}, \textit{groundbreaking}), \textbf{rigor} (\textit{rigorous}, \textit{robust}), \textbf{scale} (\textit{comprehensive}, \textit{interdisciplinary}), \textbf{utility} ( \textit{generalizable}, \textit{transformative}), \textbf{quality} (\textit{exceptional}, \textit{prestigious}), \textbf{attitude} (\textit{interesting}, \textit{remarkable}), \textbf{problem} (\textit{daunting}, \textit{ unmet}). 

As a first step, we select the \textbf{novelty} group of adjectives, which emphasizes the novelty and innovation of proposed research, an aspect of research reinforced by the academic peer-review system and the competitive funding process. This adjective group comprises 11 members: \textit{creative}, \textit{emerging}, \textit{first}, \textit{groundbreaking}, \textit{innovative}, \textit{latest}, \textit{novel}, \textit{revolutionary}, \textit{unique}, \textit{unparalleled}, \textit{unprecedented}.

To compile our dataset, we use CQPweb~\cite{hardie2012cqpweb} to search for novelty adjectives through the recent NIH corpus abstracts, covering years from 2016 to 2020. This search yielded a total of 161,469 occurrences, covering 84,299 abstracts. Due to time and resource constraints, we limit ourselves to a smaller corpus, which can be annotated and manually examined. We randomly choose 50 samples per adjective, resulting in 550 samples, covering 545 abstracts.

\section{Proposed annotation guidelines}
\label{sec:ann_guidelines}

We designed the initial annotation guidelines with the help of a linguist expert in promotional language and corpus linguistics. The guidelines comprise several sequential steps, which might require high proficiency in English. They are designed to be easy to follow, and each step provides positive and negative examples. Determining whether an adjective carries promotional intent requires considering the context in which it is used. Therefore, we chose to evaluate adjectives within their full sentential context. With this starting point, the steps of the annotation guidelines are defined as follows.

\textbf{Step 1: Value-judgment.} Determine whether the adjective implies positive value judgment. Most of the selected adjectives do; this includes priority claims (``\textit{first} method to...''). If yes, proceed to steps 2-6. If not, the adjective is deemed as \textbf{not hype}. Such cases are typically acronyms, technical/domain-specific meaning, literal meaning or part of a proper noun phrase (``\textit{first} step'', ``\textit{Creative} Scientist, Inc. (CSI)''.

\textbf{Step 2: Hyperbolic.} If the adjective is hyperbolic or exaggerated, it is deemed as \textbf{hype}. This is a relatively unambiguous category which contains, but is not limited to, a pre-determined set of adjectives: \textit{revolutionary}, \textit{unprecedented}, \textit{unparalleled}, \textit{groundbreaking}. In the great majority of cases, these adjectives are used in a promotional manner (``\textit{unparalleled} opportunity'', ``\textit{revolutionary} tool'').

\textbf{Step 3: Gratuitous.} Determine whether the adjective adds little to the proposition. If removed, and the propositional content and structural integrity of the sentence remain unchanged (typically when adjective is used in attributive relationship, e.g., "we developed \textit{innovative} technologies"), or the adjective represents a tautology and is redundant (``we \textit{discovered} a \textit{novel} gene''), the adjective is deemed as \textbf{hype}. If, however, the propositional content of the sentence would be substantially altered (``this treatment uses a \textit{novel} approach''), or justification is given within the sentence (typically when adjective is used in predicative relationship, e.g., ``the study is \textit{innovative} because no previous research...''), the adjective is \textbf{not hype}.

\textbf{Step 4: Amplified.} If the strength of the adjective is amplified through the use of modifiers, the adjective is deemed as \textbf{Hype} (``truly \textit{novel}'', ``completely \textit{unique}'', ``highly \textit{innovative}'').

\textbf{Step 5: Coordinated.} If the adjective is coordinated with one or more hype candidates, the adjective is deemed as \textbf{Hype} (e.g., ``\textit{innovative} and \textit{creative} leader''). We also term this phenomenon \textit{adjective stacking}.

\textbf{Step 6: Broader context.} When ambiguous, consider whether the sentence contains other instances of potential hype. If the overall tone of the sentence is promotional, the adjective is deemed as \textbf{hype} (e.g., ``The faculty has an \underline{outstanding} track record of \textit{creative} and \underline{innovative} research, \underline{superb} mentoring, and \underline{robust} research funding, and thus attracts \underline{outstanding} applicants.''). This step is the most subjective and potentially requires discussion and agreement of several annotators to maintain consistency of annotation.

The proposed guidelines offer some formal definitions of what constitutes hype, and can help a human annotator determine whether an adjective is hype, depending on the context. However, in some cases, the guidelines may prove insufficient and require further discussion. Detailed examples of the annotation guidelines are given in \S\ref{sec:appendix_guidelines}.

\subsection{Annotation process}
\label{sec:annotation_process}

\begin{figure}[tb]
\centering
\includegraphics[width=\linewidth]{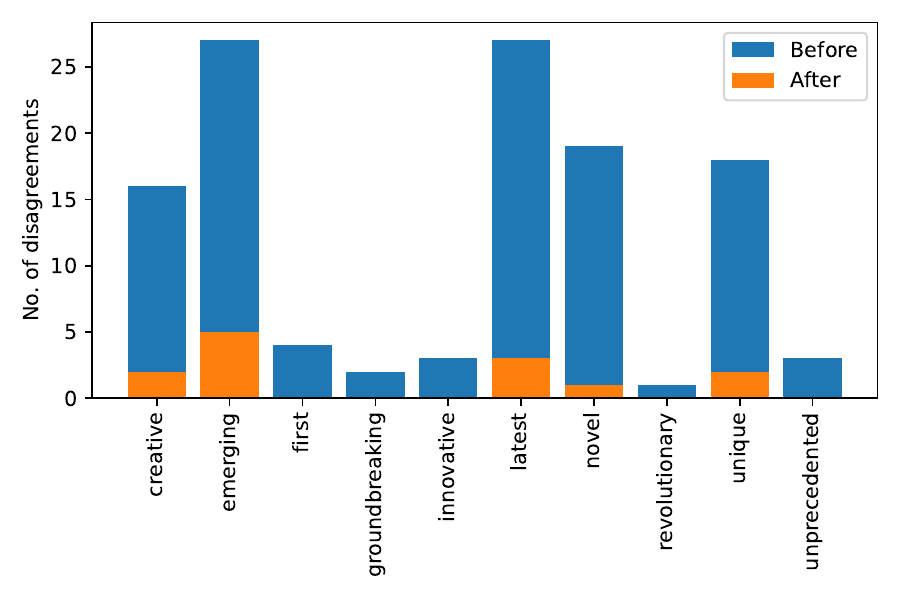}
\caption{Disagreements between annotators, before and after the discussion. Disagreements were largely resolved, except for \textit{emerging} and \textit{latest}, and to smaller extent \textit{creative}, \textit{unique} and \textit{novel}.}
\label{fig:before_and_after}
\end{figure}

Three voluntary researchers annotated the dataset, in different stages of academic career (one post-doctoral researcher, one researcher and one university Professor). They were verbally instructed to follow the proposed guidelines, knowing that the annotations will be used for scientific publication. This process followed the proposed guidelines step-by-step. In principle, an adjective can satisfy multiple criteria and be deemed \textbf{hype}; therefore, the annotators are encouraged to include every criteria that has been met. After the initial stage, a discussion session was held to resolve conflicts and reevaluate annotation guidelines. The initial stage resulted in fair amounts of disagreement (119 out of 550 samples), as indicated by the pairwise Cohen's Kappa values ranging between 0.60 and 0.78, shown in in Table~\ref{tab:cohen's}. The disagreement level differs for each adjective; as can be seen in Figure~\ref{fig:before_and_after}. 

For adjectives corresponding to the \emph{hyperbolic} guideline, the disagreements were minimal as almost all applications of the guideline result in a \textbf{hype} decision, although one annotator actively attempted to find situations where the hyperbolic usage of adjectives was justified, contrary to the guideline. This was resolved during discussion. There were no disagreements over \textit{unparalleled} - it was always judged as \textbf{hype}.

\textit{Innovative} was rarely disagreed upon (in most cases it is used in a promotional manner), as well as \textit{first}, adjective most commonly used as a numbering device (e.g., \textit{first} weeks of therapy).

Adjectives such as \textit{emerging} and \textit{latest} were the most difficult, and the guidelines proved insufficient to fully categorize them. They are often used to refer to \textit{emerging} phenomena when establishing context for the proposed research (e.g., "In the past decade, \textit{emerging} neuroimaging technique" vs. "represent a significant \textit{emerging} worldwide health threat") or the \textit{latest} publications presented at a scientific conference (e.g., "This meeting is focused on the \textit{latest} findings..." vs. "...session on applications of the \textit{latest} new technologies"); in these cases, it is necessary to look at the broader context to determine if their use is promotional.

\begin{table}[tb]
    \small
    \centering
    \tabcolsep 3pt
    \caption{Pairwise Cohen's Kappa between the annotators A, B and C. Adjusted agreement values after the discussion stage are displayed in brackets.}
    \begin{tabular}{lccc}
        \toprule
        & \textbf{A} & \textbf{B} & \textbf{C} \\
        \midrule
        \textbf{A} & -- & 0.61 (0.94) & 0.78 (0.98) \\
        \textbf{B} & 0.61 (0.94) & -- & 0.60 (0.95) \\
        \textbf{C} & 0.78 (0.98) & 0.60 (0.95) & -- \\
        \bottomrule
    \end{tabular}
    \label{tab:cohen's}
\end{table}

After the discussion stage, 106 out of the 119 disagreements were resolved, and the remaining 13 samples were discarded from the dataset. These samples highlight the difficulty of designing the guidelines, especially for adjectives \textit{emerging}, \textit{latest}, \textit{unique}, and \textit{creative}, and will be further analyzed to improve guidelines in the future.

\subsection{NIH hype dataset}

\begin{table}[tb]
    \small
    \centering
    \tabcolsep 3pt
    \caption{Final annotations for each adjective.}
    \begin{tabular}{lccc}
        \toprule
        \textbf{Adjective} & \textbf{Hype}  & \textbf{Not Hype} & \textbf{Hype \%} \\
        \midrule
        creative & 33 & 15 & 68 \\
        emerging & 22 & 23 & 48 \\
        first & 17 & 33 & 34 \\
        groundbreaking & \textit{50} & 0 & \textit{100} \\
        innovative & 41 & 9 & 82 \\
        latest & 28 & 19 & 59 \\
        novel & 18 & 31 & 36 \\
        revolutionary & \textit{50} & 0 & \textit{100} \\
        unique & 33 & 15 & 68 \\
        unparalleled & \textit{50} & 0 & \textit{100} \\
        unprecedented & \textit{50} & 0 & \textit{100} \\
        \midrule
        total  & 392 & 145 & 72\\
        \bottomrule
    \end{tabular}
    \label{tab:adjectives}
\end{table}

\begin{table*}[tb]
    \small
    \centering
    \tabcolsep 3pt
    \caption{Distribution of rationales for `hype' classification following the annotation guidelines.}
    \begin{tabular}{lccccc}
        \toprule
        \textbf{Adjective} & \textbf{Hyperbolic}  & \textbf{Gratuitous} & \textbf{Amplified} & \textbf{Coordinated} & \textbf{Broader context} \\
        \midrule
        creative        & 0     & 32    & 2     & 11    & 20    \\
        emerging        & 1     & 15    & 0     & 6     & 9     \\
        first           & 0     & 1     & 0     & 0     & 6     \\
        groundbreaking  & 50    & 2     & 1     & 5     & 12    \\
        innovative      & 2     & 39    & 3     & 4     & 18    \\
        latest          & 0     & 26    & 1     & 2     & 14    \\
        novel           & 2     & 13    & 0     & 3     & 8     \\
        revolutionary   & 50    & 1     & 1     & 2     & 16    \\
        unique          & 1     & 29    & 1     & 7     & 15    \\
        unparalleled    & 50    & 0     & 0     & 3     & 17    \\
        unprecedented   & 50    & 1     & 0     & 0     & 14    \\
        \midrule
        total           & 205   & 159   & 9     & 43    & 149   \\
        \bottomrule
    \end{tabular}
    \label{tab:rationales}
\end{table*}

The annotation process yielded a dataset of 537 sentences with potential hype adjectives. The breakdown of rationales behind the annotators' decisions, and the breakdown of \textbf{hype} and \textbf{not hype} classes can be seen in Tables~\ref{tab:adjectives} and \ref{tab:rationales}, respectively.

Table~\ref{tab:adjectives} shows that 392 adjectives are deemed as \textbf{hype}, and 145 \textbf{not hype} by the annotators. Four adjectives are \textbf{hype} in $100\%$ of the cases -- \textit{groundbreaking}, \textit{revolutionary}, \textit{unparalleled}, \textit{unprecedented}. These adjectives correspond to the `Hyperbolic' category, and are extreme examples of hype. Adjectives such as \textit{innovative}, \textit{creative} and \textit{unique} are more often than not used in promotional manner, with $82\%$, $68\%$ and $68\%$ of their uses constituting hype, respectively. Remaining adjectives are more varied in their use, and denote hype when used in a promotional context.

This distribution is reflected in annotator rationales, as shown in Table~\ref{tab:rationales}. The most numerous hype adjectives correspond to the `Hyperbolic' category, accounting for almost half of the hype examples in the dataset (205 samples). The second most prominent category is `Gratuitous', with adjectives such as \textit{innovative}, \textit{creative}, \textit{unique} most often representing this category, totaling 159 samples.  The least prominent is the `Amplified' category, with only 9 samples. This may reflect annotators' tendency to avoid overtly promotional language (``highly innovative''), since amplifying adjectives adds an additional layer of subjective evaluation. Similarly, `Coordinated' category comprises only 43 samples, as it is not very common to see adjective stacking in scientific writing. The third most common category is `Broader context', numbering 149 samples. This is a subjective category, and is based on how promotional is the overall sentence, as opposed to the literal meaning and the near context of the adjective itself.

\section{Preliminary experiments}
\label{sec:exp}

We frame hype detection as a text classification task: given a set of training sentences $S = \{s_i\}$, with known classes $C = \{\HYPE{}, \NOTHYPE{}\}$, we wish to classify an input sentence $s_q$ into one of the classes in $C$. We evaluate traditional text classification methods, pre-trained language models (PLMs) and large language models (LLMs) to assess the complexity of this problem. Finally, we obtain a preliminary human baseline, to better judge the construct validity of our annotation guidelines.

The dataset was split into development and test sets in an 8:2 ratio, stratified by class. For traditional classifiers and PLMs, hyperparameter search was done on the development set with 10-fold cross-validation, and performance on the hold-out test set is reported. LLMs were prompted whether an adjective is used in a promotional manner, given the sentence and the broad definition of hype~\cite{millar2022trends}. A verbalizer maps the spans of HYPE and NOT HYPE in the response to appropriate labels. Because we did not modify default generation settings of LLMs, the same prompt might result in different outcomes. Therefore, for each sentence, we prompted the model 5 times and report the majority decision. Example of a full prompt is given in \S\ref{sec:appendix_prompt}. Human baseline was obtained by asking a fourth voluntary researcher to manually classify the hold-out test set; we did not supply them with annotation guidelines, as we wanted to see what they would deem promotional. 

We evaluate using standard classification metrics, i.e., accuracy, precision, recall and f1-score. We look at these metrics per class and use the weighted average to account for class imbalance.

\subsection{Traditional Text Classification Methods}
\label{sec:traditional_methods}
As baseline, we train traditional text classification methods: Multinomial Naive Bayes (\MNB), Multivariate Bernoulli Naive Bayes (\MVB), Latent Semantic Analysis (\LSA), and Support Vector Machines with a linear kernel (\SVC), implemented in scikit-learn. For features we consider bag-of-words of unigrams and the averaged word embedding obtained via pre-trained \GloVe\footnote{glove.42B.300d}. To establish a lower-bound baseline, we include \Majority{}, a classifier always predicting \HYPE{}.

Results are in Table~\ref{tab:traditional_methods}. 
All traditional methods had similar performances. Interestingly, methods based on \BoW{} were still capable of performing reasonably well. However, \LSA{} was the only method performing worse than the \Majority{} baseline. With LSA, we classified sentences according to the 1-nearest neighbor strategy, curious to see if sentences with similar meanings would help identify hype. Given that it was the worst-performing method, this may indicate that hype does not occur in a specific type of text. \GloVe{} + \SVC{} led to the best performance, but it still is behind the human baseline. This result indicates hype detection requires more complex language modeling than BoW features and word embeddings. 

\begin{table}[t]
\centering
\small
\tabcolsep 3pt
\caption{Performance of traditional text classifiers in terms of Accuracy (Acc.), weighted Precision (Prec.), weighted Recall (Rec.) and weighted F1-score (F1)}
\label{tab:traditional_methods}
\begin{tabular}{llcccccc}
\toprule
\textbf{Method} & \textbf{Feature} & \textbf{Acc.} & \textbf{Pre}. & \textbf{Rec}. &\textbf{F1} \\
\midrule
\Human & - & \textbf{0.824 }& \textbf{0.819} & \textbf{0.824} & \textbf{0.821} \\
\Majority & - & 0.731 & 0.535 & 0.731 & 0.618 \\
\midrule
\MNB & \multirow{4}{*}{\BoW} & 0.741 & 0.713 & 0.741 & 0.716 \\
\MVB & & 0.741 & 0.713 & 0.741 & 0.716 \\
\LSA &  & 0.685 & 0.671 & 0.685 & 0.677 \\
\SVC &  & 0.759 & 0.736 & 0.759 & 0.717 \\
\midrule
\SVC & \GloVe & 0.796 & 0.784 & 0.796 & 0.781 \\
\bottomrule
\end{tabular}
\end{table}

\subsection{Language Models}
\label{sec:language_models}

Given the size of our dataset, we chose to finetune smaller PLMs, such as \BERT\footnote{https://huggingface.co/google-bert/bert-base-uncased} and \DistilBERT\footnote{https://huggingface.co/distilbert/distilbert-base-uncased}. We also finetuned a model pre-trained on biomedical data\footnote{https://huggingface.co/microsoft/BiomedNLP-BiomedBERT-base-uncased-abstract}, \BiomedBERT, to see if prior knowledge of the biomedical field would affect performance. Finally, we finetuned a comparable-size causal language model, \GPT\footnote{https://huggingface.co/openai-community/gpt2}. For all models, we added a classification head and performed two training schemes: 1) freezing the PLM and training only the classification head (\Snowflake), and 2) finetuning the PLM along with the classification head (+Finetuning). 

We used AdamW optimizer with HuggingFace's default hyperparameters, varying only the weight decay and learning rate through a 10-fold cross-validation on the development, with learning rate values of $[1e-5, 2e-5, 3e-5]$, weight decay values of $[0.0, 0.01, 0.1]$ and batch size of $[8, 16, 32]$. We set the number of epochs to 30, but also implemented early stopping with patience of $3$. Most models did not train past 10 epochs. The best hyperparameters found are reported in \S\ref{sec:appendix_hyperparam}. Finally, we also experimented with two LLMs, \Llama{}\footnote{https://huggingface.co/meta-llama/Llama-3.1-8B-Instruct} and \ChatGPT.

\begin{table}[t]
\centering
\small
\tabcolsep 3pt
\caption{Performance of language models in terms of Accuracy (Acc.), weighted Precision (Prec.), weighted Recall (Rec.) and weighted F1-score (F1)}
\label{tab:language_models}
\begin{tabular}{lcccccc}
\toprule
\textbf{Method} & \textbf{Acc.} & \textbf{Pre}. & \textbf{Rec}. &\textbf{F1} \\
\midrule
\Human                  & 0.824 & 0.819 & 0.824 & 0.821 \\
\midrule
\BERT \Snowflake        & 0.731 & 0.676 & 0.731 & 0.648 \\
\, +Finetuning          & \textbf{0.861} & \textbf{0.857} & \textbf{0.861} & \textbf{0.854} \\
\midrule
\DistilBERT \Snowflake  & 0.731 & 0.535 & 0.731 & 0.618 \\
\, +Finetuning          & 0.843 & 0.849 & 0.843 & 0.845 \\
\midrule
\BiomedBERT \Snowflake  & 0.695 & 0.528 & 0.695 & 0.600 \\
\, +Finetuning          & 0.843 & 0.847 & 0.843 & 0.826 \\
\midrule
\GPT \Snowflake & 0.731 & 0.535 & 0.731 & 0.618 \\
\, +Finetuning  & 0.852 & 0.848 & 0.852 & 0.844 \\
\midrule
\Llama   & & & &\\    
\, +Broad Prompt  & 0.639 & 0.716 & 0.639 & 0.659 \\
\, +Strict Prompt  & 0.657 & 0.684 & 0.657 & 0.668 \\
\midrule 
\ChatGPT & & & & \\
\, +Broad Prompt & 0.833 & 0.828 & 0.833 & 0.821 \\
\, +Strict Prompt & 0.824 & 0.840 & 0.824 & 0.796 \\
\, +Likert Prompt & 0.731 & 0.777 & 0.731 & 0.744 \\
\bottomrule
\end{tabular}
\end{table}

Results in Table~\ref{tab:language_models} show that all PLMs performed poorly when frozen, falling behind the traditional methods. Finetuning leads to substantial performance improvements, with \BERT{} outperforming all methods, including \Human{}.

Overall, we can see that \BiomedBERT{}'s prior knowledge on medical data does not seem to help in the performance, being the worst performing PLM. Furthermore, it is interesting to note that while \GPT{} performed on pair with \DistilBERT{}, it has almost the double the number of parameters. We believe that \BERT-like models can perform better with less parameters as they are bidirectional and, therefore, have a better overall understanding of the entire sentence. This result highlights the importance of the surrounding context of the potential hype words, as the rationale might be expressed before or after the word.

\Llama{} achieved the lowest performance overall; its confusion matrix shows that it often predicts \HYPE{} sentences as \NOTHYPE. This is likely due to the broad definition of hype (+Broad Prompt), under which the model can rely on inherent biases from the training data -- many promotional adjectives have become very common in the scientific parlance and, therefore, might not seem out of place. When asked to adhere more strictly to the definition of hype, the model changed its decision from \NOTHYPE{} to \HYPE{} (+Strict Prompt) for some sentences.

On the other hand, \ChatGPT{} performed on par with the human baseline. Interestingly, its best results were achieved when using the broad prompt, which is aligned with the instructions given to our human evaluator. However, prompting it with the strict prompt led to performance degradation.

\subsection{\Human{} Baseline}
Based on Tables~\ref{tab:traditional_methods} and \ref{tab:language_models}, we can see that our \Human{} baseline performed better than traditional text classifiers, but fell behind finetuned PLMs.

The confusion matrices of our \Human{} baseline and \BERT{} in Figure~\ref{fig:confusion_matrices} show that they only differ in the classification of \HYPE{} sentences, where \Human{} misclassified it more than \BERT{}. Looking closer at the eight misclassified \HYPE{} sentences, we noticed that six contained the word \textit{latest}, one contained the word \textit{emerging} and one contained the word \textit{revolutionary}.

\textit{Latest} and \textit{emerging} were the most difficult words to annotate in our dataset, as mentioned in Section~\ref{sec:annotation_process}. This result highlights the need for strict annotation guidelines. 

\subsection{Discussions}

\begin{figure}[!t]
    \centering
    \begin{subfigure}{0.75\columnwidth}
        \includegraphics[width=\columnwidth]{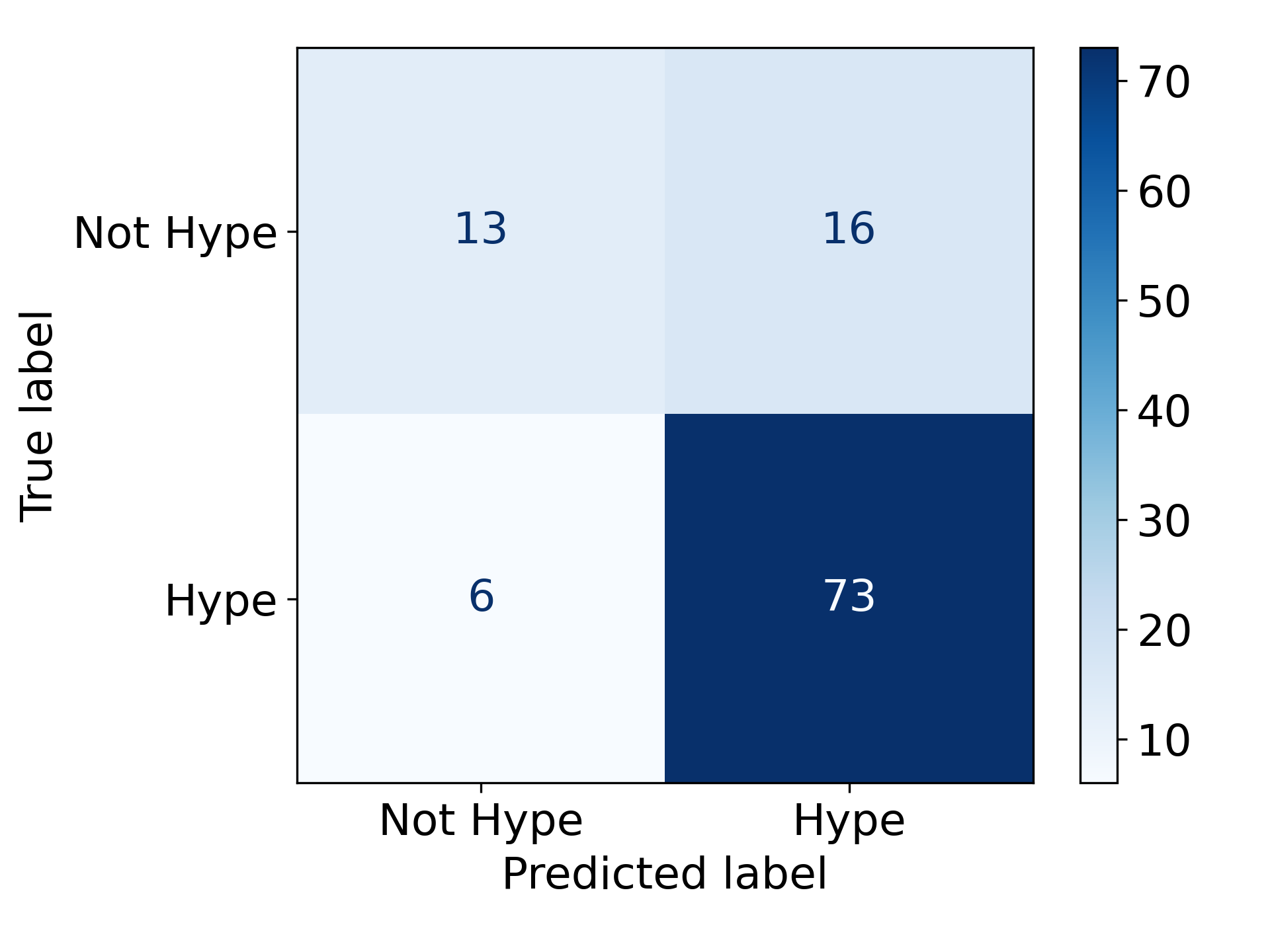}
        \caption{\GloVe{} + \SVC}
        \label{cm_glove_svm}
    \end{subfigure}
    \begin{subfigure}{0.7\columnwidth}
        \includegraphics[width=\columnwidth]{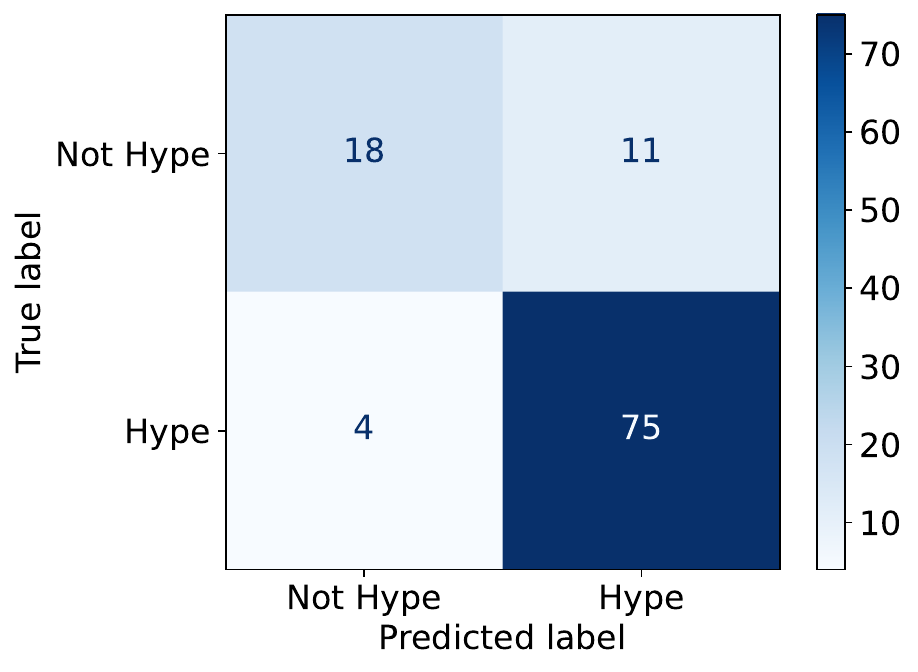}
        \caption{\BERT}
        \label{cm_bert}
    \end{subfigure}
    \begin{subfigure}{0.75\columnwidth}
        \includegraphics[width=\columnwidth]{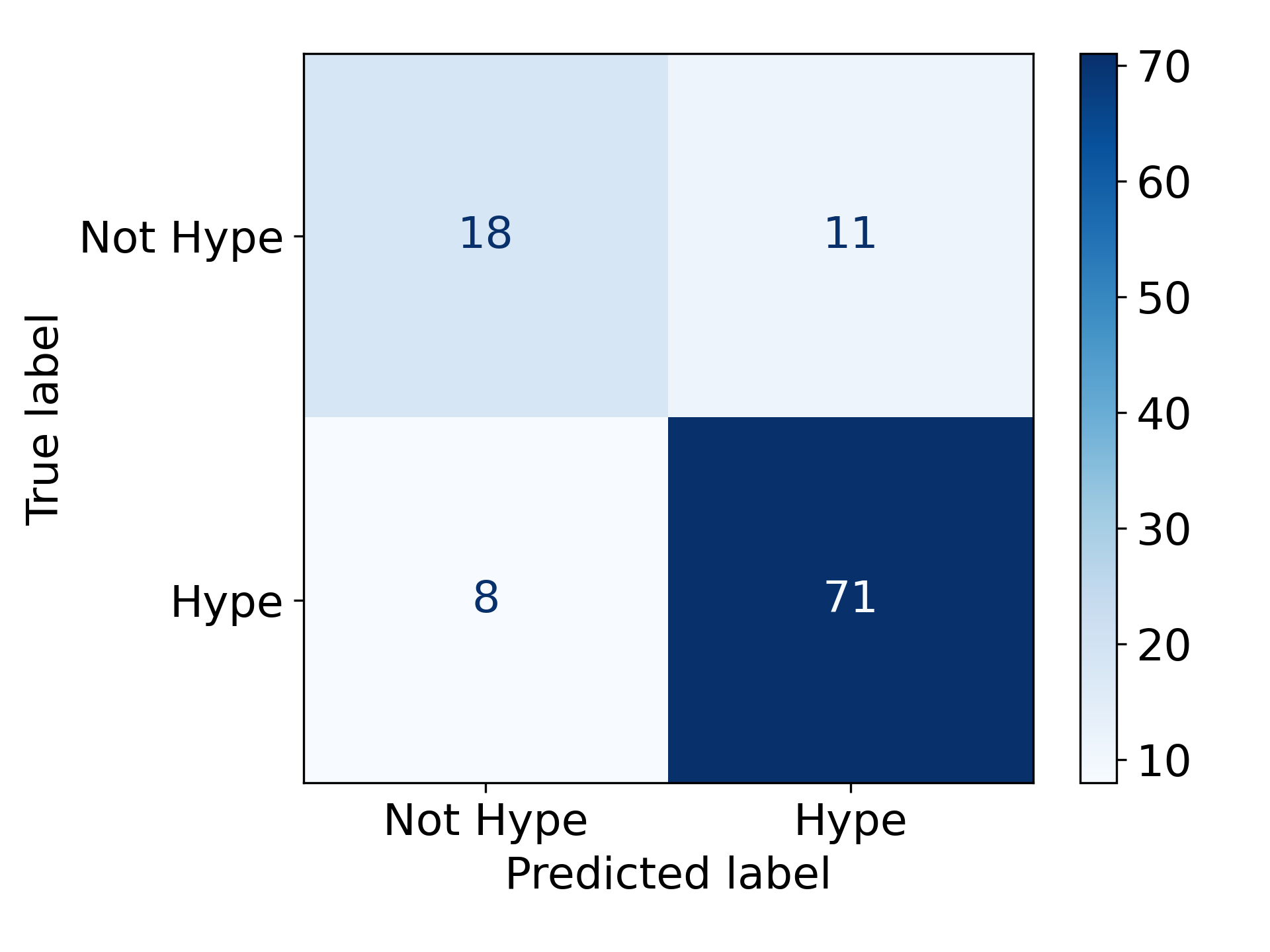}
        \caption{\Human}
        \label{fig:cm_nb}
    \end{subfigure}
    \caption{Confusion matrices for the best traditional classifier, best language model and the \Human{} baseline. While they had similar Recall on \HYPE{} sentences, \GloVe{} + \SVC{} struggles to correctly detect \NOTHYPE{} sentences, while \Human{} misclassified \HYPE{} more than \BERT.}
    \label{fig:confusion_matrices}
\end{figure}

Our experiment results have shown that the proposed guidelines align well with a blind \Human{} baseline and that language models can achieve comparable performance to this baseline. To better understand their struggles, we also analyzed their confusion matrices. For the sake of space, we show only the ones for \SVC{} + \GloVe, the best-performing language model, and the human baseline in Figure~\ref{fig:confusion_matrices}. We can see that models struggle to correctly detect \NOTHYPE{} sentences.

\begin{figure}[t]
    \centering
    \includegraphics[width=\columnwidth]{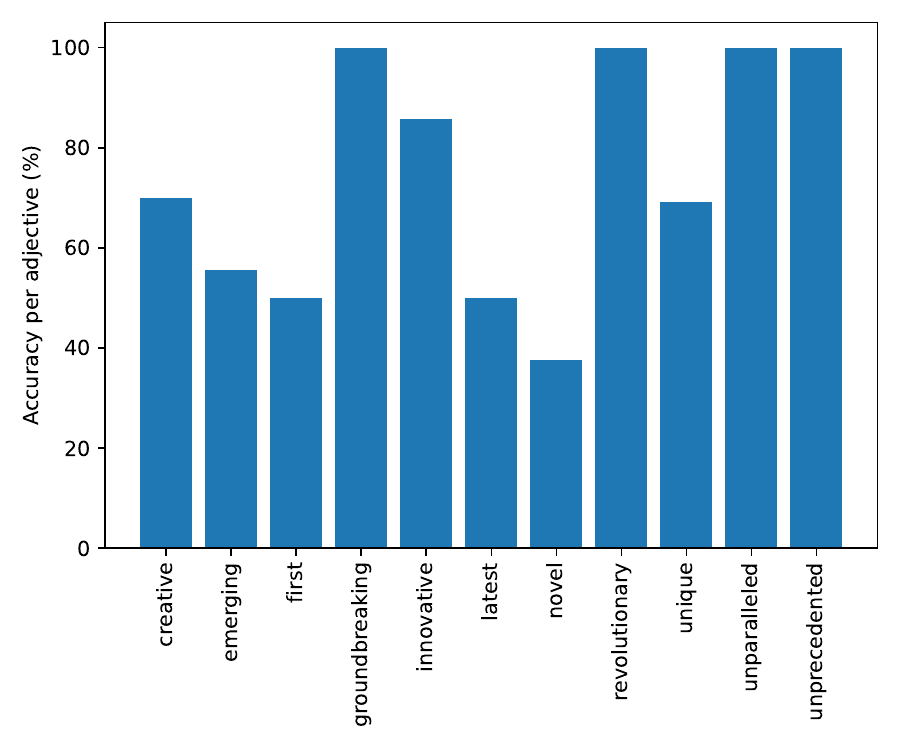}
    \caption{Accuracy per adjective according to \MNB{} + \BoW.}
    \label{fig:acc_per_adjective}
\end{figure}

Some adjectives, such as \textit{groundbreaking} and \textit{revolutionary}, only appeared in sentences labeled as \HYPE, and possibly biased the results towards only classifying sentences as \HYPE, as we can observe in Figure~\ref{fig:acc_per_adjective}. Based on our guidelines, these words fall in the `Hyperbolic' rationale and therefore were deemed as \textbf{hype} with no consideration of a broader context. 

However, our \Human{} baseline showed that hyperbolic words such as \textit{revolutionary} might not be perceived as hype. 
One might argue that stating that a method is \textit{revolutionary} does not constitute \textbf{hype} if this method was in fact \textit{revolutionary}. Alternatively, even if the method truly is revolutionary, calling it so may still be considered promotional. In a scientific context, such evaluative language can impose a judgment on the reader rather than letting the research speak for itself, thus raising questions about objectivity and, indeed, shifting values.

Now, it is important to note that one can only state if something is \textit{revolutionary} or \textit{groundbreaking} \textbf{after} observing its impact on the research field in question. Therefore, determining if these words represent hype or not requires not only a general knowledge of research field in question, but also a \textit{temporal} awareness of facts. The same applies to \textit{latest} and \textit{emerging}. 

\section{Conclusion}
\label{sec:conclusion}

To the best of our knowledge, this is the first attempt at identifying promotional language in scientific texts. We developed formal annotation guidelines and applied them to a set of texts from the NIH archive. Multiple machine learning models were used for the defined text classification task, determining whether a sentence containing a potentially promotional adjective is \HYPE{} or \NOTHYPE. Our results suggest that formalizing annotation guidelines may help humans reliably annotate adjectives as \HYPE{} or \NOTHYPE, and that using such an annotated dataset to train machine learning models yields promising results. We highlight the complexity of the task, and potential impact of domain knowledge and temporal awareness of facts.

Progress in this task requires refinement of our annotation guidelines, engineering, and better grounding in the real-world. For future work, we plan on developing guidelines for other categories of hype besides 'novelty', and greatly expanding the lexicon, possibly using automatic expansion techniques, to include adverbs, nouns and verbs, among others. We also plan on scaling up both the data annotation and human evaluation processes to better understand the perception of promotional language in scientific community. Finally, we are particularly interested in applying the current work to downstream tasks, such as automated tools for flagging and editing of promotional language in scientific writing.

\section*{Limitations}
This work proposed the task of automatically detecting hype language in biomedical research. While our proposed annotation guidelines and preliminary experiments on annotated data following these guidelines have shown promising results, there are important limitations worth noting.

First, the perception of hype is subjective. While we tried to approach this task logically, we cannot ignore our inherent biases when creating the guidelines and annotating the data.

Second, we only produced annotation guidelines for one of the eleven adjective groups within the NIH grant application abstract corpus. We expect to face similar challenges when developing guidelines for the other groups, but we cannot guarantee that we will achieve similar results.

Third, our proposed dataset is limited to approximately only 400 sentences from a corpus of almost a million abstracts. Experiments and analysis on such a small dataset might not be representative of a general distribution.

Fourth, we focus only on biomedical research. While academic writing has a distinct style, hype language in other fields is likely expressed using different words and at different intensity levels.

Fifth, we only tackled the detection of hype for English. Studying promotional language in scientific publishing in other languages might require different approaches.

Finally, experiments shown in this work were performed using a 16-GB NVIDIA V100 GPU. We reckon that experimenting with more recent LLMs will require larger computing capacity.

\section*{Ethics Statement}
Our goal is to mitigate the use of hype language in scientific publishing in biomedical research. However, one may misuse the findings of this paper to purposely include hype language.

\bibliography{acl_latex}

\newpage

\appendix
\section{Some examples in the annotation guidelines}
\label{sec:appendix_guidelines}

\emph{Guideline 1: Value-judgement.} Does the adjective imply positive value judgment?

\begin{itemize}
    \item \textbf{YES} - Most adjectives will imply a value judgement. This includes priority claims:
    \begin{itemize}
        \item \textit{Our study will be the \textbf{first} to ...}
    \end{itemize}
    \item \textbf{NO} - Typically, acronyms, technical/domain-specific meaning, and literal meaning:
    \begin{itemize}
        \item \textit{To aid these efforts, \textbf{Creative} Scientist, Inc. (CSI)...}
        \item \textit{Our curriculum emphasizes the development of critical and \textbf{creative} independent thinking...}
        \item \textit{In the \textbf{first} aim we test the hypothesis...}
    \end{itemize}
\end{itemize}

\emph{Guideline 2: Hyperbolic.} Is the adjective hyperbolic or exaggerated?

\begin{itemize}
    \item \emph{YES} - A relatively unambiguous class that can (likely) be pre-determined:
    \begin{itemize}
        \item \textit{revolutionary; unprecedented; unparalleled; groundbreaking}
    \end{itemize}
\end{itemize}

\emph{Guideline 3: Gratuitous.} Is the adjective gratuitous, adding little to the propositional content?

\begin{itemize}
    \item \emph{YES (1)} - If removed, the propositional content and structural integrity of the sentence would remain basically unchanged (typically when adjective used in attributive relationship). 
    \begin{itemize}
        \item \textit{To address this, we developed 2 \textbf{innovative} technologies.}
        \item \textit{Delivering SGR interventions via text messaging is an \textbf{innovative} way to increase the reach of this cessation intervention...}
    \end{itemize}
    \item \emph{YES (2)} - Represents a tautology or is redundant?
    \begin{itemize}
        \item \textit{discovered a \textbf{novel} gene...} 
    \end{itemize}
    \item \emph{NO (1)} - If removed the propositional content of the sentence would be substantially altered.
    \begin{itemize}
        \item \textit{This is a high risk and high impact project that uses a \textbf{novel} approach to aggressively treat local - regional disease.}

    \end{itemize}
    \item \emph{NO (2)} - The sentence gives justification for the claim (typically when adjective used in predicative relationship).
    \begin{itemize}
        \item \textit{The proposed study is \textbf{innovative} because no previous research has identified how MBC...}
    \end{itemize}
\end{itemize}

\emph{Guideline 4: Amplified.} Is the strength of the adjective amplified? 

\begin{itemize}
    \item \emph{YES} - The strength of the adjective made stronger through the use of modifiers:
    \begin{itemize}
        \item \textit{truly novel; highly innovative; completely unique; etc.}
    \end{itemize}
\end{itemize}

\emph{Guideline 5: Coordinated.} Is the adjective COORDINATED with other hype candidates?

\begin{itemize}
    \item \emph{YES} - Adjective is co-ordinated with one or more hype candidates (adjective stacking):
    \begin{itemize}
        \item \textit{...\textbf{innovative} and \textbf{creative} leader...}
        \item \textit{...\textbf{creative}, \textbf{collaborative}, and culturally \textbf{diverse} translational scientists...}
    \end{itemize}
\end{itemize}

\emph{Guideline 6: Broader context.} When ambiguous, consider whether the sentence contains other instances of potential hype or overt amplification.

\begin{itemize}
    \item \textit{This \textbf{transformative} work will be the \textbf{first} study to achieve this level of}
    \item \textit{The faculty has an \textbf{outstanding} track record of \textbf{creative} and high - profile research , \textbf{superb} mentoring , and \textbf{robust} research funding , and thus attracts \textbf{outstanding} trainees }
\end{itemize}

\section{Large Language Model Prompts}
\label{sec:appendix_prompt}

In Figures~\ref{fig:prompt_broad} and \ref{fig:prompt_strict}, we share prompts used to prompt the \Llama{} model in Section \ref{sec:language_models}, with the broad definition of hype, and its stricter variant.

\begin{figure}[h]
\begin{tcolorbox}[colback=gray!5!white,colframe=black!75!black,title=Broad Definition Prompt]
\footnotesize
You are an expert in linguistics, science communication and biomedicine. Following the DEFINITION of hype, determine whether the ADJECTIVE is used in promotional manner in a given SENTENCE. Output HYPE if yes, NOT HYPE if not. Output only the decision. Do not output reasoning. \\
    \\
    DEFINITION: If the adjective has positive value judgment, and can be removed or replaced without loss in meaning, it is potentially hype.\\
    \\
    ADJECTIVE: revolutionary\\
    \\
    SENTENCE: This research is innovative because it is the first to examine a revolutionary hypothesis on the origin of the second zygotic centriole, using a mammalian model specifically designed for this purpose.
\end{tcolorbox}
\caption{Prompt fed to \Llama{} with a broad definition of hype.}
\label{fig:prompt_broad}
\end{figure}

\begin{figure}[h]
\begin{tcolorbox}[colback=gray!5!white,colframe=black!75!black,title=Strict Definition Prompt]
\footnotesize
You are an expert in linguistics, science communication and biomedicine. Following the DEFINITION of hype, determine whether the ADJECTIVE is used in promotional manner in a given SENTENCE. Output HYPE if yes, NOT HYPE if not. Output only the decision. Do not output reasoning. \\
\\
DEFINITION: If the adjective has positive value judgment, and can be removed or replaced without loss in meaning, it is potentially hype. If the adjective can be removed without loss of meaning, it is considered HYPE.\\
\\
ADJECTIVE: revolutionary\\
\\
SENTENCE: This research is innovative because it is the first to examine a revolutionary hypothesis on the origin of the second zygotic centriole, using a mammalian model specifically designed for this purpose.
\end{tcolorbox}
\caption{Prompt fed to \Llama{} with a stricter definition of hype.}
\label{fig:prompt_strict}
\end{figure}

\section{Best Hyperparameters}
\label{sec:appendix_hyperparam}

We report the best hyperparameters found for finetuning the PLMs in Section~\ref{sec:exp} in Table~\ref{tab:hyperparam}.

\begin{table}[h!]
\centering
\small
\tabcolsep 3pt
\caption{Hyperparameters for finetuning the PLMs. Batch refers to the training batch size, WD refers to weight decay and LR refers to learning rate.}
\label{tab:hyperparam}
\begin{tabular}{lrrr}
\toprule
\textbf{Model} & \textbf{Batch} & \textbf{WD} & \textbf{LR}\\
\midrule
\BERT \Snowflake        & 8 & 0.00 & 1e-05 \\
\, +Finetuning          & 8 & 0.10 & 3e-05 \\
\DistilBERT \Snowflake  & 32 & 0.01 & 1e-05 \\
\, +Finetuning          & 8 & 0.01 & 1e-05 \\
\BiomedBERT \Snowflake  & 8 & 0.00 & 2e-05 \\
\, +Finetuning          & 16 & 0.10 & 1e-05 \\
\GPT \Snowflake         & 16 & 0.01 & 2e-05 \\
\, +Finetuning          & 8	 & 0.00	& 2e-05 \\
\bottomrule
\end{tabular}
\end{table}

\end{document}